%% file: main.tex
\documentclass[10pt,twocolumn,letterpaper]{article}

\usepackage[pagenumbers]{iccv} % To force page numbers, e.g. for an arXiv version

\usepackage{multirow}
\usepackage{amssymb} % 引入 amssymb 包 
\usepackage{colortbl}
\definecolor{lightblue}{rgb}{0.85,0.92,1.0} % 定义浅蓝色
\usepackage{marvosym}


\definecolor{iccvblue}{rgb}{0.21,0.49,0.74}
\usepackage[pagebackref,breaklinks,colorlinks,allcolors=iccvblue]{hyperref}

\def\paperID{3364} % *** Enter the Paper ID here
\def\confName{ICCV}
\def\confYear{2025}

\title{GenDoP: Auto-regressive Camera Trajectory \textbf{Gen}eration as a \textbf{D}irector \textbf{o}f \textbf{P}hotography}

\newcommand{\DatasetName}{DataDoP}
\newcommand{\MethodName}{GenDoP}

\author{Mengchen Zhang$^{1,2}$, Tong Wu$^{3}$\textsuperscript{\Letter}, Jing Tan$^{4}$, Ziwei Liu$^{5}$, Gordon Wetzstein$^{3}$,
Dahua Lin$^{2,4}$\textsuperscript{\Letter}\\
\normalsize $^{1}$Zhejiang University, 
$^{2}$Shanghai Artificial Intelligence Laboratory, 
$^{3}$Stanford University, \\
\normalsize $^{4}$The Chinese University of Hong Kong, $^{5}$Nanyang Technological University\\
\tt\small zhangmengchen@zju.edu.cn,
\{wutong16,gordon.wetzstein\}@stanford.edu\\ \tt\small
 \{tj023,dhlin\}@ie.cuhk.edu.hk, ziwei.liu@ntu.edu.sg
 }

\begin{document}

% \maketitle

\input{Figures/teaser}
\input{sec/0_abstract}    
\input{sec/1_intro}
\input{sec/2_related_work}

\input{sec/3_dataset}
\input{sec/4_method}

\input{sec/5_experiment}

\input{sec/6_conclusion}
% {
%     \small
%     \bibliographystyle{ieeenat_fullname}
%     \bibliography{main}
% }
\newpage
\clearpage

\appendix
\section*{Appendix}
% \title{GenDoP: Auto-regressive Camera Trajectory \textbf{Gen}eration as a \textbf{D}irector \textbf{o}f \textbf{P}hotography\\
% - Supplementary Materials -}

\setcounter{table}{0}
\setcounter{figure}{0}
\setcounter{footnote}{0}
\renewcommand{\thesection}{\Alph{section}}
\renewcommand\thefigure{R\arabic{figure}}
\renewcommand\thetable{S\arabic{table}}

\maketitle
\input{sec/supplementary}    
{
    \small
    \bibliographystyle{ieeenat_fullname}
    \bibliography{main}
}
\end{document}

%% file: Figures/teaser.tex
% \renewcommand\twocolumn[1][]{#1}
%
\twocolumn[{
\maketitle
\begin{center}
    \centering
    \vspace{-15pt}
    \includegraphics[width=\linewidth]{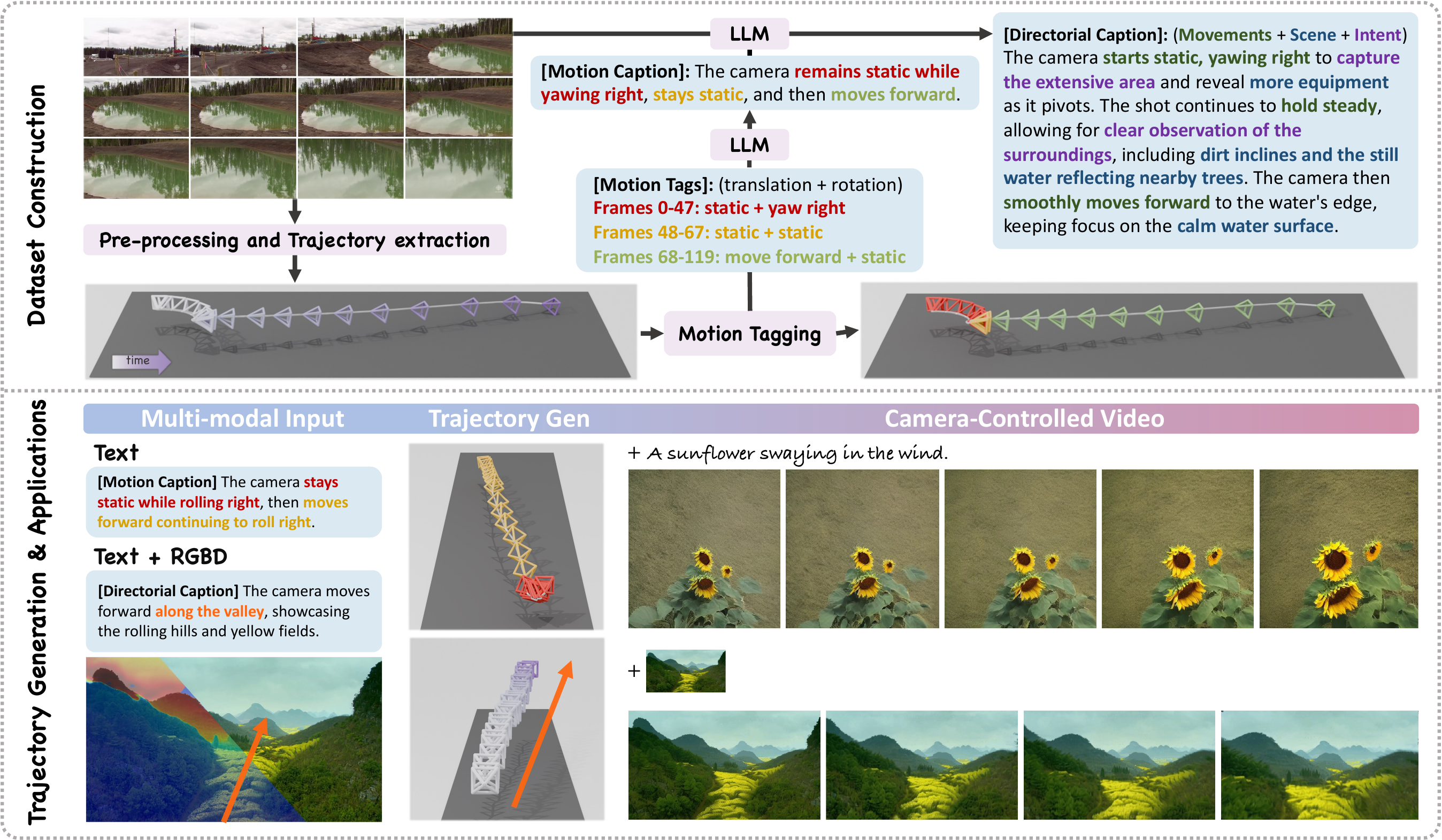}
    \setlength{\abovecaptionskip}{0mm}
    \vspace{-5pt}
    \captionof{figure}{\small \textbf{Overview.} 
    \textbf{Top}: \DatasetName{} data construction. 
Given RGB video frames, we extract RGBD images and camera poses, then tag the pose sequence with different motion categories (in different colors). With LLM, we generate two types of captions from motion tags and RGBD inputs: \textit{Motion Caption} describes the camera movements, while \textit{Directorial Caption} describes the \textcolor[rgb]{0.21875,0.33984375,0.13671875}{\textbf{camera movements}} along with their \textcolor[rgb]{0.12109375,0.3046875,0.47265625}{\textbf{interaction with the scene}} and \textcolor[rgb]{0.47265625,0.1875,0.625}{\textbf{directorial intent}}. \textbf{Bottom}: Our GenDoP method supports multi-modal inputs for trajectory creation. The generated camera sequence can be easily applied to various video generation tasks, including text-to-video (T2V)~\cite{cameractrl} and image-to-video (I2V) generation~\cite{camtrol}. GenDoP paves the way for future advancements in \textit{camera-controlled} video generation.}
\label{fig:teaser}
\end{center}}]
% \vspace{10pt}

%% file: sec/0_abstract.tex
\begin{abstract}

Camera trajectory design plays a crucial role in video production, serving as a fundamental tool for conveying directorial intent and enhancing visual storytelling. In cinematography, Directors of Photography meticulously craft camera movements to achieve expressive and intentional framing. However, existing methods for camera trajectory generation remain limited: Traditional approaches rely on geometric optimization or handcrafted procedural systems, while recent learning-based methods often inherit structural biases or lack textual alignment, constraining creative synthesis.
In this work, we introduce an auto-regressive model inspired by the expertise of Directors of Photography to generate artistic and expressive camera trajectories.
We first introduce \textbf{DataDoP}, a large-scale multi-modal dataset containing 29K real-world shots with free-moving camera trajectories, depth maps, and detailed captions in specific movements, interaction with the scene, and directorial intent.
Thanks to the comprehensive and diverse database, we further train an auto-regressive, decoder-only Transformer for high-quality, context-aware camera movement generation based on text guidance and RGBD inputs, named \textbf{GenDoP}. Extensive experiments demonstrate that compared to existing methods, GenDoP offers better controllability, finer-grained trajectory adjustments, and higher motion stability. 
We believe our approach establishes a new standard for learning-based cinematography, paving the way for future advancements in camera control and filmmaking. 
% Our code and data will be publicly available.
Our project website: \href{https://kszpxxzmc.github.io/GenDoP/}{\texttt{https://kszpxxzmc.github.io/GenDoP/}}.
\end{abstract}

%% file: sec/1_intro.tex
\section{Introduction}
\label{sec:intro}
In video production, the camera serves as the window of observation, playing a crucial role in presenting scene content, conveying the director's intent, and achieving visual effects. In recent years, video generation technology has advanced~\cite{kong2024hunyuanvideo, opensora, SVD, Cosmos}, and several cutting-edge studies have explored camera-controlled video generation~\cite{cameractrl,camtrol,dimensionx, SG-I2V}. However, these works often rely on predefined, simplistic camera trajectories to demonstrate their results. The generation of artistic, expressive, and intentional camera movements remains largely unexplored.

Trajectory generation has been a long-standing problem.
Traditional approaches include optimization-based camera motion planning~\cite{DBLP:journals/cga/Blinn88a, DBLP:journals/tog/LinoC15, DBLP:conf/mig/GalvaneCLR15} and learning-based camera control~\cite{DBLP:journals/jfr/BonattiWHAGCKCS20, DBLP:conf/si3d/DruckerGZ92, DBLP:conf/cvpr/HuangLYK00C19, DBLP:journals/tog/JiangWWCC20}.
However, these techniques demand geometric modeling or cost function engineering for each motion, which limits creative synthesis. Meanwhile, oversimplified procedural systems impede precise text control.
Recent advances in diffusion-based camera trajectory generation~\cite{CCD,ET, Director3D} have expanded creative possibilities for text-driven cinematography. However, CCD~\cite{CCD} and E.T.~\cite{ET} inherit structural biases from human-centric tracking datasets, constraining camera movements to oversimplified character-relative motion patterns. Director3D~\cite{Director3D} introduces object/scene-centric 3D trajectories from multi-view datasets~\cite{MVImgNet, RealEstate10k}, but the lack of trajectory-level captions limits text-to-motion alignment. As a result, the generated paths are driven by geometric plausibility rather than directorial intent.
These dataset constraints hinder the creation of artistically coherent free-moving trajectories that interpret creative vision without relying on specific subjects.

In this work, we tackle the problems above with several key designs.
First, we introduce \textbf{\DatasetName{}} Dataset, a multi-modal, free-moving camera motion dataset extracted from real video clips, which includes accurate camera trajectories extracted by state-of-the-art and scene compositions.
We extract camera trajectories and corresponding depth maps using MonST3R~\cite{MonST3R}, and employ GPT-4o to generate comprehensive descriptions of the camera trajectories and scene focus, capturing both motion dynamics and directorial intent.
\DatasetName{} comprises over 29K shots, totaling 11M frames, with corresponding camera trajectories and diverse textual descriptions. 
Furthermore, given the inherently sequential nature of camera trajectories, we propose\textbf{ \MethodName{}}, which treats camera parameters as discrete tokens and leverages an auto-regressive model for camera trajectory generation. Our model incorporates multi-modal condition as inputs, including fine-grained textual descriptions and optionally RGBD information from the first frame, to produce stable, complex, and instruction-aligned camera movements.

We conduct rigorous human validation to ensure the dataset quality.  Extensive experiments confirm that \MethodName{} outperforms state-of-the-art methods~\cite{CCD, ET, Director3D} across fine-grained textual controllability, motion stability, and complexity, while exhibiting enhanced robustness.
As AI-driven video creation evolves, multi-modal camera trajectory generation emerges as a timely and crucial direction. We believe that this work paves the way for future advancements in camera-controlled video generation and a wide range of trajectory-related downstream applications.

%% file: sec/2_related_work.tex
\input{Tables/dataset}

\section{Related Work}
\label{sec:related_work}

\noindent\textbf{Camera trajectory datasets.}
While several existing datasets document camera trajectories, their cinematographic expressiveness remains constrained. 
Datasets such as MVImgNet~\cite{MVImgNet}, RealEstate10K~\cite{RealEstate10k}, and DL3DV-10K~\cite{DL3DV-10K} provide calibrated trajectories through structured capture methods, but predominantly focus on basic paths around static objects or scenes. These datasets lack the sophisticated cinematographic language necessary for narrative-driven sequencing and intentional viewpoint control. 
CCD~\cite{CCD} and E.T.~\cite{ET} emphasize human-centric tracking but are confined to reactive tracking mechanisms. 
In contrast, \DatasetName{}'s camera movement is driven by the compositional logic of the scene and the narrative demands. We underscore \DatasetName{}'s unique contribution to the field of artistic camera trajectory generation.

\noindent\textbf{Camera trajectory generation.}
Early efforts in trajectory generation generally consist optimization-based motion planning~\cite{DBLP:journals/cga/Blinn88a, DBLP:journals/tog/LinoC15, DBLP:conf/mig/GalvaneCLR15,SplaTraj} and learning-based camera control~\cite{DBLP:journals/jfr/BonattiWHAGCKCS20, DBLP:conf/si3d/DruckerGZ92, DBLP:conf/cvpr/HuangLYK00C19, DBLP:journals/tog/JiangWWCC20}.
Recent progress focuses on integrating camera motion with scene and character dynamics. CCD~\cite{CCD} introduced a camera diffusion model using text and keyframe controls, generating motion in character-centric coordinates. E.T.~\cite{ET} improves to incorporate both character trajectories and camera-character text descriptions as control and generates trajectories in the global coordinates. On the other hand, Director3D~\cite{Director3D} trains DiT-based framework on object/scene-level multi-view datasets to generate object/scene-centric camera trajectories. NWM~\cite{NWM} employs conditional DiT to plan camera trajectory via agents' egocentric views. 
Concurrent work~\cite{dvgformer} employs an auto-regressive transformer to predict the next frame’s camera movement based on past camera paths and images in aerial videography. Our approach goes further by incorporating both text instructions and RGBD spatial information, enabling precise control in generating camera trajectories for cinematic storytelling.

\noindent\textbf{Auto-regressive models.}
Auto-regressive (AR) modeling employs tokenizers to transform inputs into discrete tokens and formulates generation as a next-token prediction task with transformers. 
In recent years, great advancements are witnessed in auto-regressive modeling in image~\cite{DBLP:conf/icml/RameshPGGVRCS21,DBLP:conf/nips/TianJYPW24,DBLP:journals/tmlr/YuXKLBWVKYAHHPLZBW22,Infinity}, video~\cite{VideoGPT,VideoPoet,Loong}, and 3D generation~\cite{EdgeRunner, MeshAnything, MeshGPT}.
Early approaches~\cite{DBLP:conf/icml/RameshPGGVRCS21,DBLP:journals/tmlr/YuXKLBWVKYAHHPLZBW22} serialize images into patch tokens and train a transformer to auto-regressively model the text and image tokens in a sequential data stream. VAR~\cite{DBLP:conf/nips/TianJYPW24} reformulates auto-regressive image generation as coarse-to-fine next-scale prediction. 
VideoPoet\cite{VideoPoet} leverages bidirectional attention for multi-modal input conditioning in auto-regressive video generation. 
Our work extends auto-regressive modeling to camera trajectory generation controlled by text and geometry cues, leveraging the discrete nature of camera tokens. Compared to diffusion-based methods, our model generates precise, coherent, and intricately detailed artistic trajectories for long camera pose sequences.

%% file: Tables/dataset.tex
\begin{table*}
% \small
% \scriptsize
\centering
\vspace{-15pt}
\setlength{\tabcolsep}{8pt}
\renewcommand{\arraystretch}{0.8}
\fontsize{8}{10.5}\selectfont
\begin{tabular}{l|cc|cccc|ccc}
\toprule
\multirow{2}{*}{\textbf{Dataset}} & \multirow{2}{*}{\textbf{Traj Type}} & \multirow{2}{*}{\textbf{Domain}} & \multicolumn{4}{c|}{\textbf{Caption}} &  \multicolumn{3}{c}{\textbf{Statistics}} \\
 & & & \textbf{Traj} & \textbf{Scene} & \textbf{Intent} & \textbf{\#Vocab} & \textbf{\#Sample} & \textbf{\#Frame} & \textbf{\#Avg (s)} \\
\midrule
MVImgNet~\cite{MVImgNet} & Object/Scene-Centric & Captured & $\times$ & $\times$ & $\times$ & - &  22K & 6.5M & 10 \\
RealEstate10k~\cite{RealEstate10k} & Object/Scene-Centric & Youtube & $\times$ & $\times$ & $\times$ & - &  \textbf{79K} & 11M & 5.5 \\
DL3DV-10K~\cite{DL3DV-10K} & Object/Scene-Centric & Captured & $\times$ & $\times$ & $\times$ & - &  10K & \textbf{51M} & \textbf{85} \\
\midrule
CCD~\cite{CCD} & Tracking & Synthetic & \textbf{$\checkmark $} & $\times$ & $\times$ & 48 &  25K & 4.5M & 7.2 \\
E.T.~\cite{ET} & Tracking & \textbf{Film} & \textbf{$\checkmark $} & $\times$ & $\times$ & 1790 &  \textbf{115K} & \textbf{11M} & 3.8 \\
\DatasetName{} (Ours) & \textbf{Free-Moving} & \textbf{Film} & \textbf{$\checkmark $} & \textbf{$\checkmark $} & \textbf{$\checkmark $} & \textbf{8698} &  29K & \textbf{11M} & \textbf{14.4} \\
\bottomrule
\end{tabular}
\vspace{-6pt}
\caption{\textbf{\DatasetName{} Dataset.} We compare the \DatasetName{} dataset to other datasets containing camera trajectories. \DatasetName{} is a large dataset focusing on artistic, free-moving trajectories, each accompanied by high-quality caption annotations. The provided captions detail the camera movements, their interactions with scene content, and the underlying directorial intent. 
To capture more intricate camera movements, each video clip spans 10-20 seconds, averaging 14.4 seconds.}
\vspace{-8pt}
\label{tab:dataset}
\end{table*}

%% file: sec/3_dataset.tex
\section{\DatasetName{} Dataset}
\label{sec:dataset}

We introduce \textbf{\DatasetName{}}, a camera trajectory dataset extracted from long shots in artistic films, including both movies and documentaries, designed to capture free-moving, intricate, and expressive camera movements. As shown in \cref{fig:teaser}, each sample in \DatasetName{} consists of a shot-level camera trajectory, accompanied by the corresponding RGBD images and two types of trajectory captions: \textbf{Motion} captions, which accurately describe the camera motion alone, and \textbf{Directorial} captions, which detail the camera movements, their interaction with the scene, and the directorial intent. We describe the data construction pipeline in \cref{sec:Construction} and the dataset statistics in \cref{sec:Statistics}.

\subsection{Dataset Construction}
\label{sec:Construction}
%  dataset展示图
% 1. Data Collection 
% 2. Data Process
%     Boundary Detection
%     Remove Caption 
%     Shot Filter 
%         black ratio 
%         Statics and Motion
% 3. Pose Extraction(Monst3r) Monst3r的参数
% 4. Description GPT4的prompt提示部分+最终得到的文本描述

% The \DatasetName{} construction pipeline, illustrated in \TODO{Figure 3: pipeline}, is described in detail in the following steps.

\input{Figures/statistics}

\noindent\textbf{Pre-processing.}
We curate and filter artistic videos from the internet, which are then segmented into shots using PySceneDetect~\footnote{https://github.com/Breakthrough/PySceneDetect}. Captions are removed using VSR~\footnote{https://github.com/YaoFANGUK/video-subtitle-remover}, after which the shots are merged with a publicly available subset from MovieShots~\cite{MovieShots}. A filtering process is applied to retain shots between 10 and 20 seconds in length, while removing those that are excessively dark or nearly static. Since our dataset focuses on free-moving camera trajectories, which enable unrestricted 3D camera motion within scenes and events, rather than tracking moving people or objects, we specifically filter for this category of data. GPT-4o~\cite{GPT4O} was used to categorize the shots, removing those with static cameras or object-tracking motion. For details, please refer to~\cref{sec:Construction_sup}.

\noindent\textbf{Trajectory extraction.}
We then utilize MonST3R~\cite{MonST3R} to estimate the geometry of dynamic scenes. Camera trajectories are extracted along with the corresponding depth maps. The trajectories are subsequently cleaned, smoothed, and interpolated into fixed-length sequences.

\noindent\textbf{Motion tagging.}
We then partition the camera trajectories into segments of motion tags. Compared to existing datasets~\cite{CCD, ET}, our captions explicitly incorporate descriptions of camera rotation, enabling more fine-grained characterization of camera movements. As a result, our motion tags include both translation and rotation components (see~\cref{fig:trans_rot}). For camera translation, excluding the static state, we consider six fundamental motions across three degrees of freedom: lateral (left/static/right), vertical (up/static/down), and depth (forward/static/backward). Each translation motion can be categorized into one, two, or three motions, resulting in a total of 27 possible combinations. For camera rotation, aside from the static state, we consider six fundamental motions across three degrees of freedom: pitch (up/down), yaw (left/right), roll(left/right), resulting in 7 base actions. We do not consider the combination of these rotations, as in practical scenarios, rotation typically involves only one of these basic motions at a time. We simplify by assuming that camera translation and rotation are completely independent, which results in a total of $27\times 7$ possible combinations for camera motion tags.
% \TODO{add some more explanation of the assumptions of 7 }

We adopt the motion tagging method from E.T.~\cite{ET} to process the camera trajectories. For translation, we use an initial velocity threshold and velocity difference thresholds in different directions to determine the dominant velocity direction combinations. For rotation, we use an initial rotational velocity threshold to identify the unique dominant rotational direction. Finally, we combine the translation and rotation information to generate the complete tags, and apply smoothing to remove noise and sparse tags. These methods provide a coarse temporal description of the camera trajectories, as shown in ~\cref{fig:teaser}.

\input{Tables/dataset_study}

\noindent\textbf{Caption generation.}
Finally, we generate two types of trajectory captions based on the motion tags obtained in the previous stage, as shown in~\cref{fig:teaser}. First, we structure the motion tags by incorporating context, instructions, constraints, and examples, and then leverage GPT-4o to generate \textbf{Motion} captions that describe the camera motion alone. Next, we extract 16 evenly spaced frames from the shots to create a $4\times4$ grid and prompt GPT-4o to consider both the previous caption and the image sequence. This enables GPT-4o to generate \textbf{Directorial} captions that describe the camera movement, the interaction between the camera and the scene, as well as the directorial intent. Further details can be found in~\cref{sec:Construction_sup}.
% Appendix Sec. A.2.

% \tong{Need to align the names: we have ``pure'' vs. ``mixed'' at the first paragraph, and we have ``tag'' vs ``caption'' here.}

\subsection{Dataset Statistics}
\label{sec:Statistics}
% 【插表与其他dataset的对比】

\noindent\textbf{Trajectory types.} 
% \noindent\textbf{Scenario and Trajectory types.} 
We classify camera trajectories into four types: \textit{Static}, \textit{Object/Scene-Centric}, \textit{Tracking}, and \textit{Free-Moving}. Static shots keep the camera fixed. Object/Scene-Centric shots capture multi-view data focusing on specific objects or scenes. Tracking shots track a moving subject. Free-Moving shots allow unrestricted 3D camera motion, enabling complex scene exploration and dynamic framing, crucial for cinematic storytelling and creative expression.
As shown in \cref{tab:dataset}, \DatasetName{} stands out by uniquely focusing on artistic, free-moving trajectories, capturing the director's creative vision and offering significant cinematic and artistic value.
Unlike tracking shots, where the camera follows a specific object, free-moving shots fluidly navigate the scene, enhancing visual storytelling without constraints.

% \tong{Free-tracking seems to be similar with the RE10K and DL3DV-10K, maybe we need to highlight some other good properties beyond the pure trajectory types: we contain diverse scene contents, dynamic objects for the frames. It's not good that we cannot directly use them in our current framework, maybe mention some further usages in future work section.}

% \tong{For table 1, compress the number dims where we do not look good, add some other propertie measures: diversity (scene, traj), caption (motion tag, intention， etc)}

\noindent\textbf{Data scale.}
\DatasetName{} is built on long shots from the Internet. As shown in \cref{tab:dataset}, it consists of 29K samples, spanning 12M frames and totaling 113 hours of footage, all with high-quality trajectory annotations. The dataset focuses on long shots averaging 14.4 seconds, capturing more complex camera movements compared to other datasets. While DL3DV-10K~\cite{DL3DV-10K} has a longer average duration, its camera trajectories lack directorial intent, emphasizing scene-level consistency rather than creative camera work.

\input{Figures/method}
\noindent\textbf{Statistics.}
% \TODO{Figure 2: Dataset Statistics}
% \TODO{Table 3: Dataset Quality}
We present the dataset statistics across four dimensions: \textit{Alignment}, \textit{Quality}, \textit{Complexity}, and \textit{Diversity}. 
To evaluate Alignment and Quality, we conducted a user study with 8 experts. We selected 100 samples, including original videos, camera trajectories, and two captions: Motion and Directorial. The samples were split into two sets, each labeled by four users. For \textit{Alignment}, we assess the consistency between the trajectory and video (Video-Traj), the motion caption and trajectory (Traj-Motion), and the directorial caption with both the trajectory and video scene (Traj-Directorial).
For \textit{Quality}, we assess whether the camera trajectory is free of breaks, roughness, or jitter. 
% We use Average Pairwise Agreement(APA)~\cite{APA} to compute the agreement among multiple users, ensuring the reliability of the user study. 
We use Fleiss' Kappa~\cite{kappa} to measure inter-rater agreement among multiple users.
As shown in~\cref{tab:dataset_study}, our dataset achieves high accuracy in both Alignment and Quality, with all Kappa values exceeding 0.4, confirming the reliability of the results. 
For \textit{Complexity}, as illustrated in~\cref{fig:trans_rot}, we present the composition and distribution of motion tags within the dataset.
For \textit{Diversity}, as shown in~\cref{fig:diversity}, the trajectories, while remaining consistent with the caption, exhibit significant variations in length, direction, and speed, effectively showcasing the diversity within our dataset.
% we randomly sampled 200 trajectories and applied the canonical normalization method outlined in~\cref{sec:tokenization}. The resulting trajectories, depicted in~\cref{fig:diversity}, exhibit omnidirectional distribution patterns, showcasing the diversity of camera trajectories in our dataset.

% user-study: Alignment, Quality(smooth, extraction from video)
% Pics: gpt4-Diversity get scene dis, get motion dis

% 【插图Dataset Statistics】
% 移动 / 旋转 分布 
% duration + length

%% file: Figures/statistics.tex
\begin{figure*}
  \centering
  \vspace{-15pt}
  \begin{subfigure}{0.66\linewidth}
\includegraphics[width=1\linewidth]{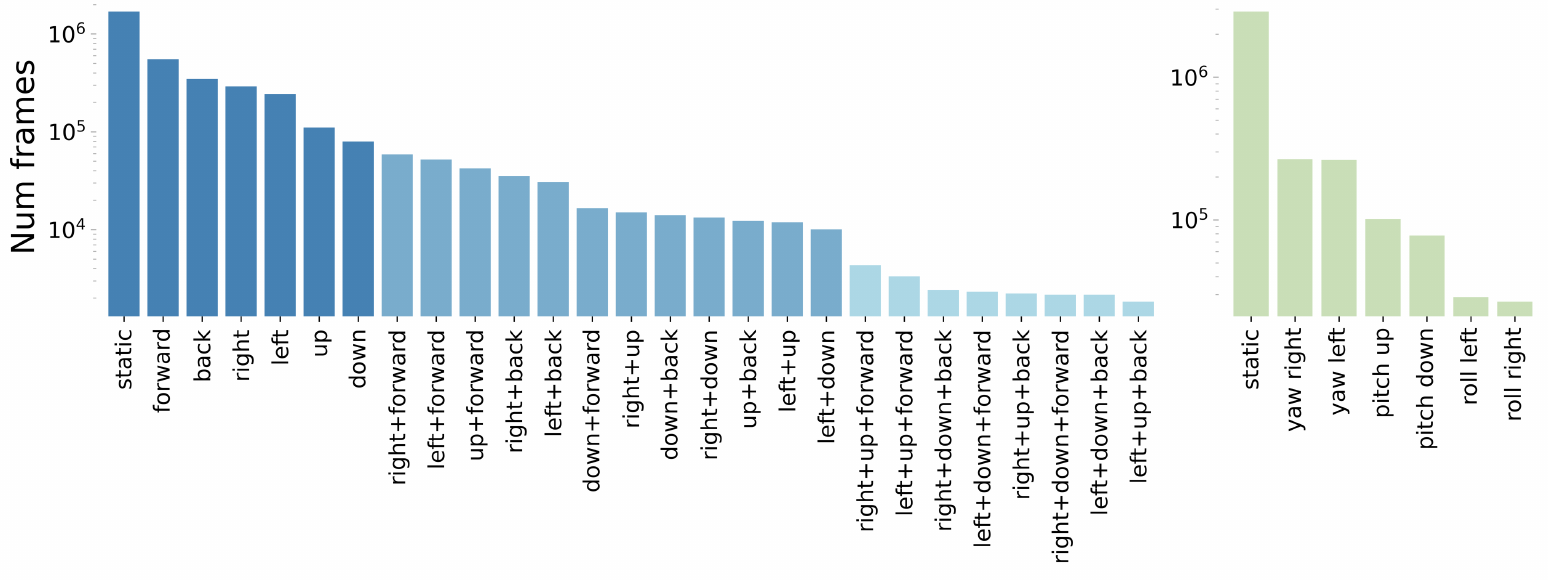}
    \vspace{-15pt}
    \caption{Distribution of Translation and Rotation Motion Tags.}
    \label{fig:trans_rot}
  \end{subfigure}
    \hfill
    \begin{subfigure}{0.31\linewidth}
    \includegraphics[width=1\linewidth]{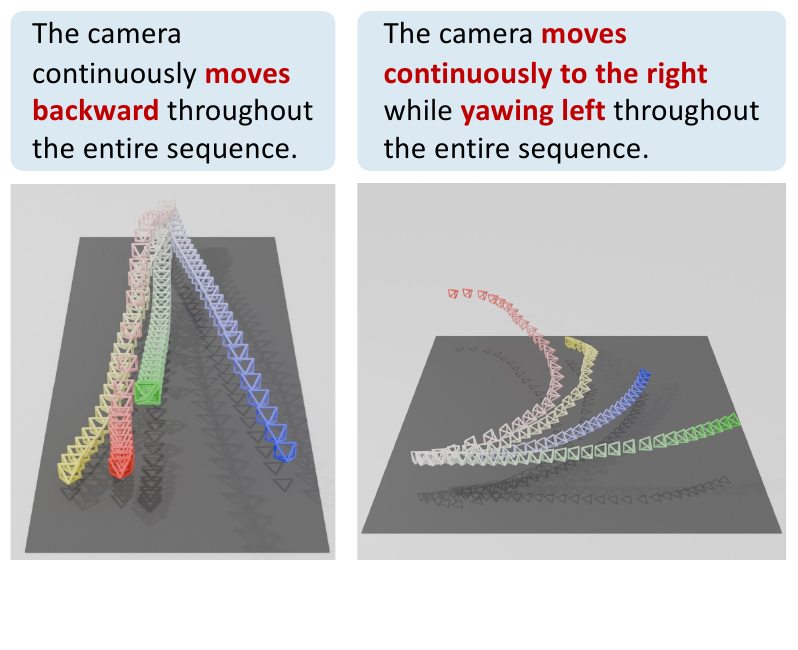}
    \vspace{-15pt}
    \caption{Diverse Trajectories.}
    \label{fig:diversity}
  \end{subfigure}
  \vspace{-9pt}
 \caption{\textbf{Dataset Statistics.} \textbf{(a)} The figure illustrates the composition and distribution of 27 translation motions (left) and 7 rotation motions (right), emphasizing the complexity and diversity of trajectories in our DataDoP dataset.
\textbf{(b)} Based on the same caption, our dataset includes diverse trajectories that still conform to the given caption. As shown in the figure, the trajectories exhibit variations in terms of length, direction, and speed, effectively showcasing the diversity within our dataset.}
  \vspace{-6pt}
  \label{fig:Statistics}
\end{figure*}

%% file: Tables/dataset_study.tex
\begin{table}%
\small
\centering
\setlength{\tabcolsep}{6pt}
    \renewcommand{\arraystretch}{0.8}
\scalebox{0.9}{\begin{tabular}{l|ccc|c}
\toprule
\multirow{2}{*}{\textbf{Score}} & \multicolumn{3}{c|}{\textbf{Alignment}} & \multirow{2}{*}{\textbf{Quality}} \\
& \textbf{Video-Traj} & \textbf{Traj-Motion} & \textbf{Traj-Directorial} &                         \\
\midrule
Acc   &0.863	&0.913	&0.858	&0.945 \\
Kappa &0.642	&0.530	&0.502	&0.551 \\
\bottomrule
\end{tabular}}
\vspace{-8pt}
\caption{\small \textbf{Dataset User Study.} Our user study demonstrates that our dataset exhibits excellent quality and human-alignment, with proven reliability of the results.}
\vspace{-15pt}
\label{tab:dataset_study}
\end{table}

%% file: Figures/method.tex
\begin{figure*}
  \centering
  \vspace{-15pt}
    \includegraphics[width=1\linewidth]{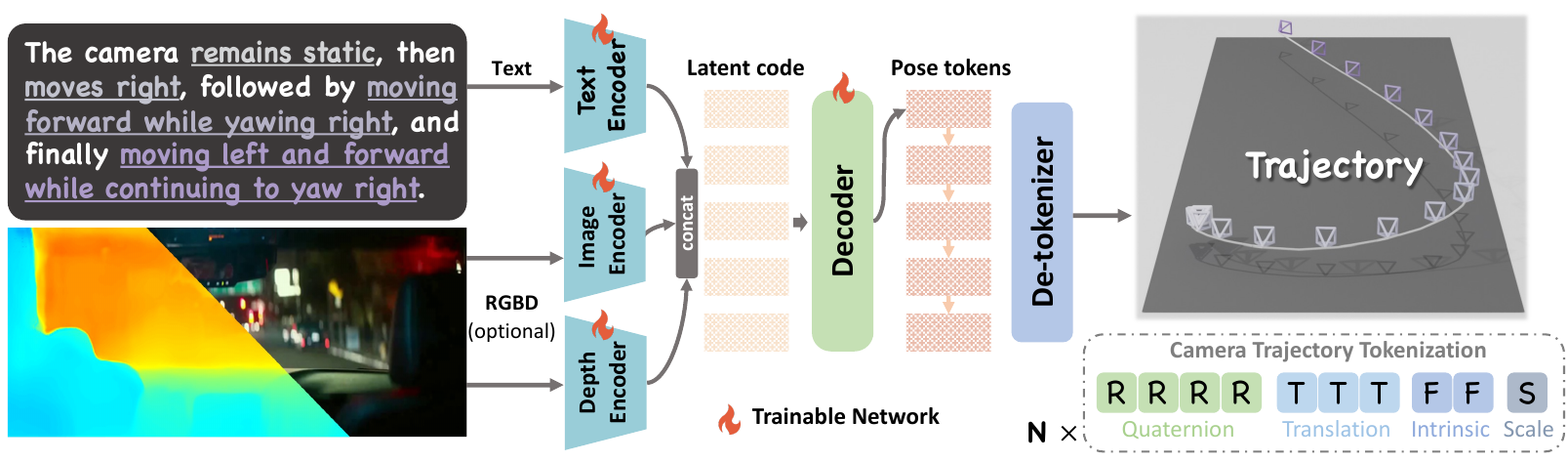}
    \vspace{-20pt}
  \caption{\textbf{Our Auto-regressive Generation Model.} Our model supports multi-modal inputs and generates trajectories based on these inputs. By treating the task as an auto-regressive next-token prediction problem, the model sequentially generates trajectories, with each new pose prediction influenced by previous camera states and input conditions.}
  \vspace{-5pt}
  \label{fig:method}
\end{figure*}

%% file: sec/4_method.tex
\section{Method}
\label{sec:method}
\subsection{Overview}
We introduce \textbf{\MethodName{}} here, an auto-regressive method for camera trajectory generation. Previous trajectory generation methods~\cite{EdgeRunner, CCD, Director3D, MDM, DTG, NWM} largely relied on diffusion models~\cite{Diffusion}, which often result in discontinuous and unstable trajectories (See~\cref{fig:qualitative}). 
In contrast, we pioneer the application of auto-regressive models to trajectory generation. Auto-regressive models are well-suited for this task due to their ability to capture sequential dependencies. In trajectory, each pose's position and orientation depend on the previous one, making the framework ideal for modeling the temporal and spatial continuity of trajectories. By conditioning each pose on its predecessor, the model effectively generates realistic and coherent 3D camera trajectories.

\MethodName{} automatically constructs the camera's 3D motion path based on an input caption or appearance and geometry from the initial frame, capturing changes in both position and orientation. As illustrated in~\cref{fig:method}, \MethodName{} takes a text description $T$, optionally combined with the initial frame's RGBD image $(I_0, D_0)$, as input and generates the corresponding camera trajectory $\mathcal{C}$. 
A camera trajectory $\mathcal{C} = \{\mathbf{x}_0, \mathbf{x}_1, \dots, \mathbf{x}_{N-1}\}$ is defined as a sequence of $N$ consecutive camera poses, where each pose $\mathbf{x}_i = [\mathbf{R}_i | \mathbf{t}_i | \mathbf{K}_i]$ comprises a rotation matrix $\mathbf{R}_i$ (orientation), a translation vector $\mathbf{t}_i$ (position), and an intrinsic matrix $\mathbf{K}_i$ (projection parameters). The intrinsic matrix $\mathbf{}{K}_i$ can be simplified to $(f_x, f_y)$, assuming a fixed principal point $(c_x, c_y)$ and image dimensions $(H, W)$. Our goal is to derive an auto-regressive generation function $f$ such that $\mathcal{C} = f(T[, (I_0, D_0)])$. The trajectory tokenization process is detailed in~\cref{sec:tokenization}, while the generation method is comprehensively described in~\cref{sec:network}.

\input{Tables/benchmark}

\subsection{Camera Trajectory Tokenization}
\label{sec:tokenization}
% 解释Tokenization化的方法。
Auto-regressive models commonly process information as discrete token sequences, making compact tokenization essential for efficient representation without sacrificing accuracy. 
Videos are naturally serialized into discrete frames and in this sense, camera trajectories from videos can be easily tokenized into discrete camera pose $\mathbf{x}_i = [\mathbf{R}_i | \mathbf{t}_i | \mathbf{K}_i]$ at each frame. This simplicity facilitates efficient tokenization, enabling a compact encoding of the trajectory. 

\noindent\textbf{Canonical normalization.} 
We first establish a scale-invariant trajectory representation via canonical normalization. The camera frame is aligned as the world reference, setting $\mathbf{R}_0^{\mathrm{norm}} = \mathbf{I}$ and $\mathbf{t}_0^{\mathrm{norm}} = \mathbf{0}$. Subsequent poses are relativized through rigid transformation: $\mathbf{R}_i^{\mathrm{norm}} = \mathbf{R}_0^\top \mathbf{R}_i$, $\hat{\mathbf{t}}_i = \mathbf{R}_0^\top (\mathbf{t}_i - \mathbf{t}_0)$ for $i \in [1, N)$. Scale normalization then computes $s = \max_{1 \leq i < N} \|\hat{\mathbf{t}}_i\|_2$ and projects translations to unit space via $\mathbf{t}_i^{\mathrm{norm}} = \hat{\mathbf{t}}_i/(s + \epsilon)$ with $\epsilon=10^{-5}$, maintaining geometric consistency and numerical stability.

\noindent\textbf{Trajectory tokenization.} 
For the resulting normalized parameters $\mathbf{R}_i^{\mathrm{norm}}$ and $\mathbf{t}_i^{\mathrm{norm}}$, we compute the corresponding quaternion representation for $\mathbf{R}_i^{\mathrm{norm}}$ and normalize all parameters to the range $[0, 1]$, resulting in the vector $(r_1, r_2, r_3, r_4, t_1, t_2, t_3)$. Subsequently, the focal lengths $f_x, f_y$ and the scale size $s$ are also normalized, yielding $(f_1, f_2, s)$, which are then concatenated with the previously computed values. Finally, these parameters are multiplied by the discrete bin size $B$ and converted into integer values.
Thus, for each $\mathbf{x}_i$, we can tokenize it into an integer vector of length 10, where the values are within the range $[0, B]$. As a result, each camera trajectory can be tokenized into an integer vector of length $10N$. 

\noindent\textbf{Auxiliary tokens.} Similar to prior auto-regressive approaches~\cite{EdgeRunner, MeshGPT, MeshAnything}, we prepend a \textbf{BOS} token at the beginning of a trajectory sequence, append an \textbf{EOS} token at the end, and use \textbf{PAD} tokens to fill the necessary positions. During the tokenization process, we obtain $B+1$ integer values. Consequently, this tokenized representation can be discretized through a learnable codebook $\mathcal{V}\in \mathcal{R}^{(B+4)\times L}$,  where $L$ is the latent dimension.

\subsection{Auto-regressive Generation}
\label{sec:network}
We employ a transformer-based auto-regressive architecture to establish a bidirectional mapping between fixed-length camera trajectories and their compact latent representations. Although raw camera trajectories may vary in length, their spatial paths remain consistent after interpolation, allowing us to target fixed-length camera trajectories.

\noindent\textbf{Text-conditioned encoder.}
Our architecture comprises a text encoder \( \mathcal{E}_T \) and an auto-regressive decoder \( \mathcal{D} \) for the base text-conditioned model, as shown in~\cref{fig:method}. The text encoder \( \mathcal{E}_T \) utilizes the pretrained and learnable text encoder from Stable Diffusion 2.1 (SD2.1)~\cite{Diffusion} to extract semantic features, which are then processed through an MLP to generate a textual latent code \( \mathbf{Z}_T \in \mathbb{R}^{M_T \times L} \), where \( M_T \) denotes the textual latent size and \( L \) is the latent dimension.

\noindent\textbf{RGBD-conditioned encoder.}
For the RGBD-conditioned model, we introduce two separate encoders: \( \mathcal{E}_I \) for RGB image and \( \mathcal{E}_D \) for depth. Specifically, we expand the depth to $\mathbb{R}^{3\times H \times W}$ to ensure it can be processed by the encoder. Both encoders use the pretrained and learnable CLIP Vision Model~\cite{LAION5B, CLIP, OpenCLIP} to extract features, which are then passed through MLPs to generate latent codes \( \mathbf{Z}_I \in \mathbb{R}^{M_I \times L} \) and \( \mathbf{Z}_D \in \mathbb{R}^{M_D \times L} \). The final latent representation is the concatenation of the textual, RGB, and depth codes:
\begin{equation}
\mathbf{Z} = [\mathbf{Z}_T; \mathbf{Z}_I; \mathbf{Z}_D] \in \mathbb{R}^{M \times L}, M = M_T + M_I + M_D.
\end{equation}
This combined representation integrates both visual and geometric modalities, conditioning the trajectory generation on the accompanying textual information.

\noindent\textbf{Auto-regressive decoder.}  
The decoder \( \mathcal{D} \) is an auto-regressive transformer designed to generate a trajectory token sequence from the latent code \( \mathbf{Z} \) and previously token IDs. We adopt the OPT architecture~\cite{OPT} as the decoder, as utilized in prior works~\cite{EdgeRunner, MeshAnything}.  
The latent code \( \mathbf{Z} \) is prepended to the input sequence, positioned before the \textbf{BOS} token. For each token prediction, the decoder \( \mathcal{D} \) queries the learnable codebook \( \mathcal{V} \in \mathbb{R}^{(B+4) \times L} \) using the previous token IDs \( \mathbf{y}_{0:P-1} \), producing the corresponding continuous token embeddings \( V[\mathbf{y}_{0:P-1}] \in \mathbb{R}^{P \times L} \), where \( P \) denotes the length of the previous token sequence. The input embeddings for the decoder are then computed as:
\begin{equation}
    \mathbf{X}_P = \mathrm{PosEmbed} ([\mathbf{Z}; V[\mathbf{y}_{0:P-1}]]) \in \mathbb{R}^{(M+P) \times L}.
\end{equation}
Stacked causal self-attention layers are then employed to predict the next feature based on \( \mathbf{X}_P \). A linear projection is applied to map the predicted feature to classification logits, which are subsequently used to retrieve the corresponding token ID \( \mathbf{y}_P \). This process ultimately generates a fixed-length trajectory token sequence.

\noindent\textbf{Loss function.}  
The model is optimized with a weighted sum of cross-entropy loss and a regularization term:
\begin{equation}
L = \mathrm{CrossEntropy}(S[1:], \hat{S}[:, -1]) + \lambda \lVert \mathbf{Z} \rVert_2^2,
\end{equation}
where \( S \) is the one-hot ground truth token sequence, \( \hat{S} \) is the predicted logits, and \( \mathbf{Z} \) is the latent code.

%% file: Tables/benchmark.tex
\begin{table*}%
% \small
\scriptsize
\centering
\vspace{-10pt}
\begin{tabular}{l|ll|cc|cc|ccc}
\toprule
\multirow{2}{*}{\textbf{Condition}} & \multirow{2}{*}{\textbf{Method}} & \multirow{2}{*}{\textbf{Dataset}} & \multicolumn{2}{c|}{\textbf{Text-Trajectory Alignment}} & \multicolumn{2}{c|}{\textbf{Trajectory Quality}} & \multicolumn{3}{c}{\textbf{User Study (AUR)}}                     \\
                                    &                                  &                                   & \textbf{F1-Score}$\uparrow$        & \textbf{CLaTr-CLIP}$\uparrow$       & \textbf{Coverage}$\uparrow$      & \textbf{CLaTr-FID}$\downarrow$     & \textbf{Alignment}$\uparrow$   & \textbf{Quality}$\uparrow$   & \textbf{Complexity}$\uparrow$   \\
\midrule
\multirow{5}{*}{\textbf{Motion}}      & CCD~\cite{CCD}                              & Pre-trained                       & 0.297                    & 5.288                       & 0.332                  & 357.822                &   3.013&	3.022&3.273                  \\
                                    & E.T.~\cite{ET}                                  & Pre-trained                       & 0.330                    & 2.450                       & 0.020                  & 609.906               &    1.227&	1.067	&1.067  \\
                                    & Director3D~\cite{Director3D}                          & Pre-trained                       & 0.058                    & 0.000                       & 0.171                  & 542.385                &                  2.313&	3.110&	2.453                       \\
                                    & Director3D~\cite{Director3D}                          & \DatasetName{}                           & \underline{ 0.391}              & \underline{ 31.689}                & \underline{ 0.839}            & \underline{ 31.979}           & \underline{3.753}&	\underline{3.260}&	\underline{3.493}                    \\
                                   \rowcolor{lightblue} \cellcolor{white}   &  \textbf{\MethodName{}(Ours)}                & \DatasetName{}                           & \textbf{0.400}           & \textbf{36.179}             & \textbf{0.872}         & \textbf{22.714}        &                   \textbf{4.693}	&\textbf{4.573}&	\textbf{4.713 }                \\
\midrule
\multirow{5}{*}{\textbf{Directorial}}     & CCD~\cite{CCD}                                & Pre-trained                       & 0.315                    & 4.247                       & 0.416                  & 240.216                &            2.950	&3.050	&3.217              \\
                                    & E.T.~\cite{ET}                               & Pre-trained                       & 0.319                    & 0.000                       & 0.014                  & 758.923                &       1.309&	1.092	& 1.184                  \\
                                    & Director3D~\cite{Director3D}                       & Pre-trained                       & 0.126                    & 0.000                       & 0.348                  & 348.312                &               2.333&	2.867&	2.342                     \\
                                    & Director3D~\cite{Director3D}                          & \DatasetName{}                           & \underline{ 0.361}              & \underline{ 23.505}                & \underline{ 0.802}            & \underline{ 35.538}           &              \underline{ 3.808}&	\underline{3.467}&	\underline{3.683 }         \\
                                  \rowcolor{lightblue}  \cellcolor{white} & \textbf{\MethodName{}(Ours)}              & \DatasetName{}                           & \textbf{0.399}           & \textbf{32.408}             & \textbf{0.854}         & \textbf{34.275}        &                 \textbf{4.617}	&\textbf{4.557}&	\textbf{4.575}                     \\
\midrule
 \rowcolor{lightblue}  \cellcolor{white} \textbf{RGBD \& Text}                & \textbf{\MethodName{} (Ours)}                & \DatasetName{}                           & \textbf{0.388}                    & \textbf{30.231}                      & \textbf{0.855}                  & \textbf{33.653}                 &         -           &             -     &       -     \\
\bottomrule
\end{tabular}
\vspace{-3pt}
\caption{\textbf{Quantitative Results.} We present the quantitative results of our GenDoP across two text-conditional generation tasks and an RGBD \& Text-conditioned task, comparing it with human-tracking methods CCD~\cite{CCD} and E.T.~\cite{ET}, as well as the object/scene-centric method Director3D~\cite{Director3D}. Our model consistently outperforms all baselines across all metrics and caption subsets, confirming the effectiveness of both our dataset and auto-regressive framework, positioning \MethodName{} as a state-of-the-art trajectory generation model.}
% \vspace{-10pt}
\label{tab:benchmark}
\end{table*}

%% file: sec/5_experiment.tex
\input{Figures/qualitative}

\section{Experiments}
\label{sec:experiment}

\subsection{Experimental Setting}
Our \MethodName{} framework implements three conditional generation paradigms: (1) \textbf{Motion} captions for isolated camera motions, (2) \textbf{Directorial} captions for scene-synchronized trajectories, and (3) \textbf{RGBD \& Text}, a novel approach that integrates images and depth maps with \textit{Directorial} captions through hierarchical feature fusion.

All experiments for both training and inference are carried out with an Intel(R) Xeon(R) Gold 6248R CPU @ 3.00GHz and a single NVIDIA A100-SXM4-80GB GPU.
We maintain consistency in parameters and strategies throughout training to ensure uniformity across the experimental setup. The image resolution is set to \( W = 512 \), with a trajectory length of \( N = 60 \), discrete bin size \( B = 256 \), and latent dimension \( L = 1024 \). The textual latent size is \( M_T = 77 \), the image latent size is \( M_V = 257 \), and the depth latent size is \( M_D = 257 \). The backbone OPT Transformer consists of 12 layers with 12 attention heads each. Training converges after 8 hours on a single A100 GPU, yielding an inference throughput of approximately 3 seconds per trajectory. For evaluation, 3k samples are randomly selected from the \DatasetName{} dataset as the test set, with the remaining data forming the training corpus. Implementation specifics are detailed in~\cref{sec:Experiments_sup}.
% Appendix Sec. C.1.

\subsection{Quantitative Results}
\noindent\textbf{Metrics.}
We obtain the Contrastive Language-Trajectory embedding (CLaTr)~\cite{ET} by leveraging the \DatasetName{} dataset with a CLIP-like approach~\cite{CLIP}. A random subset of 3k samples is selected as a test set, from which CLaTr embeddings for both ground truth (GT) and generated data are extracted. Using these embeddings, we evaluate the model with two main metrics. (1) \textbf{Text-Trajectory Alignment}: We measure the similarity between text and trajectory embeddings using the \texttt{CLaTr-CLIP} (analogous to CLIP-Score~\cite{OpenCLIP}). We also replicate the motion tagging step from~\cref{sec:Construction} to obtain the GT motion tags, which are then compared with the generated tags. Classifier \texttt{F1-Score} is computed by verifying the generated motion tags against the GT labels.
(2) \textbf{Trajectory Quality}: The alignment between GT and generated trajectories is evaluated using \texttt{CLaTr-FID} (analogous to FID~\cite{FID}). Additionally, 
\texttt{Coverage} evaluates how well the generated data spans the range of real data, with higher values indicating a broader representation of the data distribution. 

\noindent\textbf{Results.} We report the quantitative results of our \MethodName{} across two text-conditional generation paradigms (Motion / Directorial) in~\cref{tab:benchmark}. We compare it with previous trajectory generation methods. For the human-tracking methods, CCD~\cite{CCD} and E.T.~\cite{ET}, we assume the character remains static to simplify the camera trajectory inference process. For the object/scene-centric, text-only conditioned method, Director3D~\cite{Director3D}, in addition to the pretrained model, we also train a version using \DatasetName{} to emphasize the significance and effectiveness of our dataset for camera trajectory generation tasks. For the RGBD \& Text-conditioned task, a novel paradigm introduced by us, we present only the metric results for \MethodName{}.

Our model demonstrates consistent superiority across all metrics and caption subsets, primarily due to the enhanced trajectory complexity and trajectory-aware captions in our dataset.
This innovation enables more precise motion representation, significantly enhancing text-trajectory alignment. This is demonstrated by Director3D models trained on our dataset, which show a dramatic leap in CLaTr-CLIP scores from 0 to over 30, transitioning from object-centric to trajectory-enriched training.
Despite sharing the same training data, \MethodName{} outperforms \DatasetName{}-trained Director3D by 4.5 (Motion) and 9.1 (Directorial) for CLaTr-CLIP, while reducing CLaTr-FID by 9.3 (Motion) and 1.3 (Directorial), as confirmed by user studies. These results validate the effectiveness of our auto-regressive framework.
Additionally, \MethodName{} demonstrates exceptional versatility in handling RGBD \& Text-conditional tasks, showing strong multi-modal integration for high-quality trajectory generation under complex constraints. Collectively, the experiments confirm the effectiveness of both our dataset and auto-regressive framework, establishing \MethodName{} as a state-of-the-art model for trajectory generation.

\input{Figures/RGBD}

\noindent\textbf{User study.} 
To establish human-aligned evaluation metrics, we engaged 27 domain experts in a user study centered on three critical dimensions: \texttt{Alignment} (trajectory consistency with input text), \texttt{Quality} (smoothness, logical coherence, and seamless connectivity between sequential actions), and \texttt{Complexity} (kinematic sophistication of motion sequences under input constraints).
We employed the Average User Ranking (AUR) metric to evaluate model performance, where domain experts assigned ranking scores (1-5) to the five competing models per task. Higher ranking scores indicate superior performance. 
We comparatively assessed five models on text-conditioned tasks (with 10 samples per task), excluding RGBD \& Text-conditional scenarios with single-model baselines. As evidenced in~\cref{tab:benchmark}, our approach outperformed others across all metrics, with results closely matching the earlier quantitative findings, validating its perceptual and technical coherence.

\input{Tables/ablation}

\subsection{Qualitative Results}
\noindent\textbf{Text-conditioned Generation.} We present comparative analysis of Text-conditioned trajectory generation in~\cref{fig:qualitative}. Our model not only achieves superior text-trajectory alignment but also maintains high-quality trajectory generation. Furthermore, the intricate input conditions highlight its capacity to produce sophisticated outputs with high-level complexity. In contrast, the DataDoP-trained Director3D captures basic motion patterns but exhibits trajectory jitter and instability. Furthermore, its object-centric variant pre-trained on~\cite{MVImgNet,RealEstate10k} generates orbit-dominated trajectories that exhibit no text correspondence, despite improved smoothness. Other baselines exhibit notably inferior performance in both text-trajectory alignment and quality.

\noindent\textbf{RGBD \& Text-conditioned Generation.}
We conduct a comparative analysis of our trajectory generation model under varying input conditions, as shown in~\cref{fig:RGBD}. The results demonstrate that both models generate command-compliant trajectories when given identical textual inputs. However, the RGBD \& Text-conditioned model shows superior scene adaptation by leveraging RGBD to incorporate geometric and contextual constraints. Specifically, as shown in the first row of~\cref{fig:RGBD}, the spatial information from RGBD effectively mitigates ambiguities, i.e., ``left and forward'' in the text. This multimodal conditioning enables precise alignment with the 3D scene structure.

\subsection{Ablation Studies}

% \TODO{Table3: Ablation}
\noindent\textbf{Canonical normalization.}
We experiment with an alternative strategy that skips canonical normalization, directly using trajectories from Monst3r~\cite{MonST3R} with scale normalization for tokenization feasibility. These trajectories are scene-centered, with the 3D space focused around the scene. In contrast, canonical normalization transforms them into first-person tracking paths. As shown in the table~\cref{tab:ablation}, applying canonical normalization significantly improves both alignment and quality, providing more consistent camera movements align with the instructions.

\noindent\textbf{Trainable Encoder.}
Contrary to conventional practice in text/image-conditional generation where pretrained encoders remain frozen to preserve prior knowledge, our experiments demonstrate comprehensive performance gains (see~\cref{tab:ablation}) by employing trainable encoders. This improvement arises from the encoders' ability to adapt and bridge cross-modal gaps: through joint optimization, the visual encoder creates geometry-aware trajectory embeddings, while the text encoder learns motion-semantic relationships, resulting in more accurate alignment between text and camera movements.

%% file: Figures/qualitative.tex
\begin{figure*}
  \centering
  \vspace{-10pt}
    \includegraphics[width=1\linewidth]{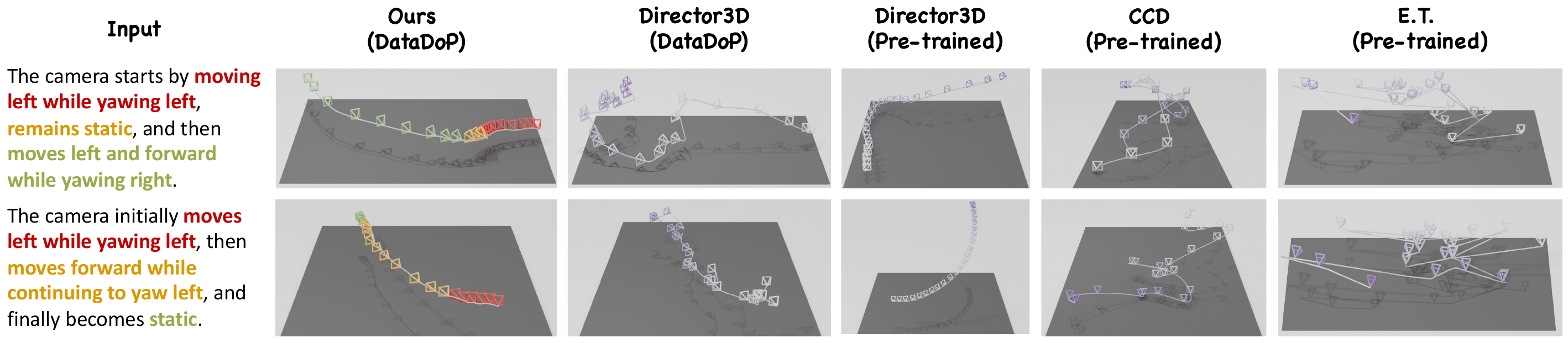}
    \vspace{-15pt}
\caption{\textbf{Qualitative Results of Text-conditioned Trajectory Generation.} We offer a comparative analysis of text-conditioned trajectory generation in the figure. Our model's trajectories (color-coded to highlight text alignment) remain stable and closely follow the instructions, while other models exhibit significant jitter or fail to match the instructions well.}
    \vspace{-5pt}
  \label{fig:qualitative}
\end{figure*}

%% file: Figures/RGBD.tex
\begin{figure}
  \centering
    \includegraphics[width=1\linewidth]{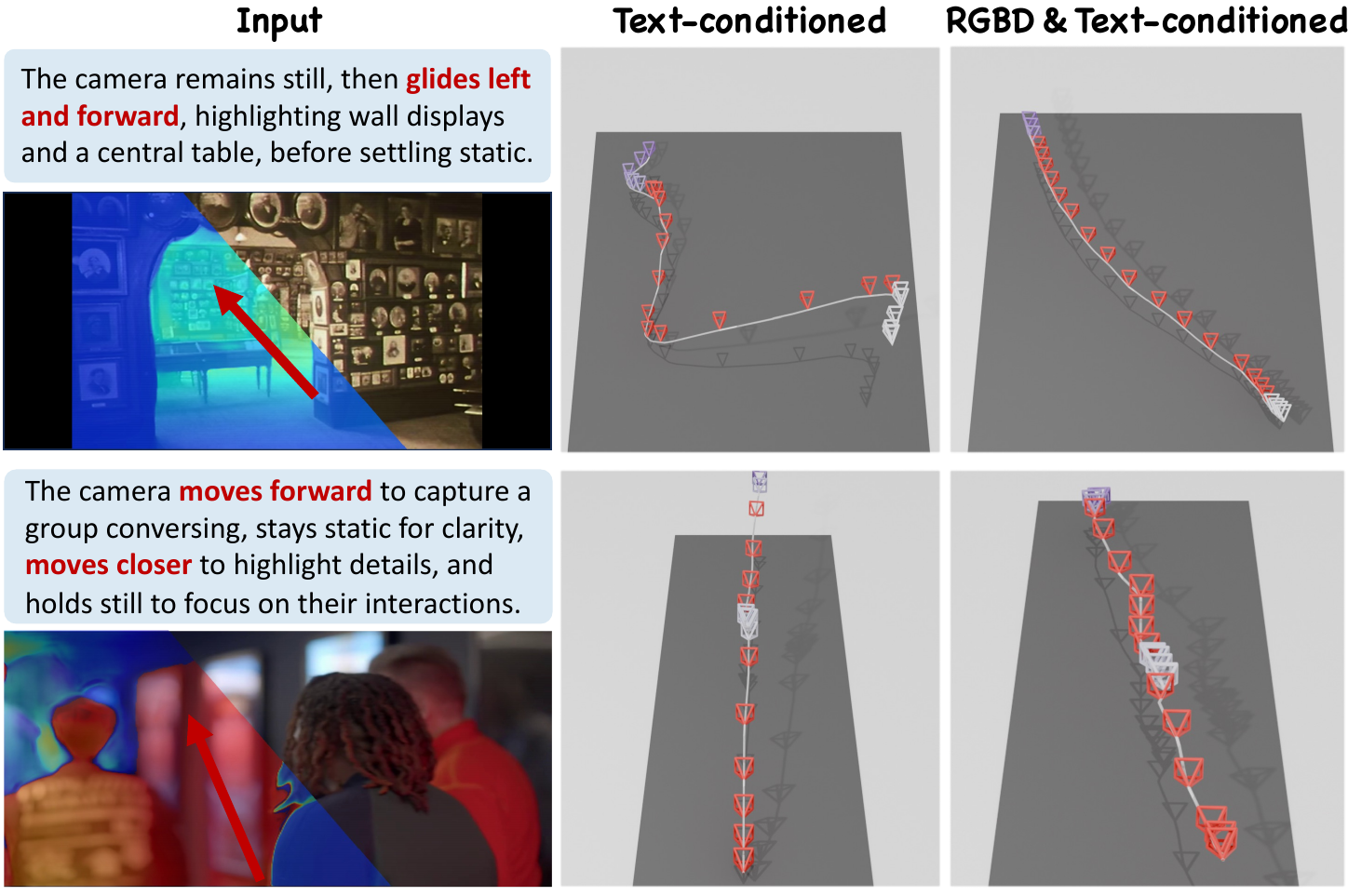}
    \vspace{-10pt}
  \caption{\textbf{Qualitative Results of RGBD \& Text-conditioned Generation.} This figure compares the impact of incorporating RGBD input on trajectory generation under identical text conditions. While both models generate command-compliant trajectories, the RGBD \& Text-conditioned model demonstrates superior scene adaptation by utilizing RGBD data to integrate geometric and contextual constraints.}
  \label{fig:RGBD}
  \vspace{0pt}
\end{figure}

%% file: Tables/ablation.tex
\begin{table}
% \small
% \scriptsize
\centering
\fontsize{6.8}{9}\selectfont
\begin{tabular}{cc|cc|cc}
\toprule
\multicolumn{2}{c}{\textbf{Ablation}}                         & \multicolumn{2}{|c}{\textbf{Text-Traj Alignment}}                           & \multicolumn{2}{|c}{\textbf{Trajectory Quality}}                                \\
\textbf{Encoder} & \textbf{Norm} & \multicolumn{1}{c}{\textbf{F1-Score}} & \multicolumn{1}{c}{\textbf{CLaTr-CLIP}} & \multicolumn{1}{|c}{\textbf{Coverage}} & \multicolumn{1}{c}{\textbf{CLaTr-FID}} \\
\midrule
$\checkmark$                          & $\checkmark$                                & \textbf{0.400}                        & \textbf{36.179}                          & \textbf{0.872}                        & \textbf{22.714}                             \\
$\checkmark$                          & $\times$                                & 0.322                     & 14.917                                   & 0.766                                 & 68.590     \\   
$\times$                          & $\checkmark$                                & 0.389                                 & 31.420                                   & 0.866                                 & 22.841                        \\

\bottomrule
\end{tabular}
\vspace{-5pt}
\caption{\textbf{Ablation Study.} We conduct an ablation study to evaluate the effectiveness of canonical normalization (see~\cref{sec:tokenization}) and the trainability of the encoder (see~\cref{sec:network}).}
\vspace{-5pt}
\label{tab:ablation}
\end{table}

%% file: sec/6_conclusion.tex
\section{Conclusion}
\label{sec:conclusion}
We propose \textbf{\DatasetName{}}, a pioneering dataset of expressive, free-moving camera trajectories from artistic videos, and \textbf{\MethodName{}}, an auto-regressive multimodal model for trajectory generation. Our approach innovatively incorporates RGBD information as input, enabling spatial data to guide trajectory supervision. This sets a new benchmark, achieving state-of-the-art performance with superior controllability and intent alignment compared to existing methods.

\noindent\textbf{Limitations and future work.} 
Currently, our multimodal approach combines text and first-frame RGBD to generate trajectories. Meanwhile, our dataset also extracts 4D point cloud during the extraction process but remains underexplored. 
Looking ahead, we aim to incorporate more modalities to enhance the adaptability and contextual awareness of the generated trajectories. 
In addition, we plan to unify trajectory and camera-controlled video creation for iterative creation of both trajectories and video content, establishing a seamless pipeline for automated, artistic film production.

%% file: sec/supplementary.tex
The appendix provides detailed supplementary material on the \textbf{DataDoP} dataset and \textbf{GenDoP} method. It outlines data availability, ensuring compliance with YouTube's policies and detailing our data sharing practices. The dataset construction process is described in detail, including shot collection, quality filtering, and semantic categorization using GPT-4o~\cite{GPT4O}. Additionally, we provide further dataset statistics. The appendix also explains the tokenization details of camera trajectory data for model processing. Finally, it includes information on the experimental setup, along with additional ablation studies

\section{DataDoP Dataset}
\label{sec:dataset_sup}
\subsection{Data Availability Statement and Clarification}
We are dedicated to upholding transparency and compliance in our data collection and sharing practices. Please take note of the following:

\begin{itemize}
\item \textbf{Publicly Available Data:} The data utilized in our studies is sourced from publicly available repositories. We do not access any exclusive or private data sources.

\item \textbf{Data Sharing Policy:} Our data sharing policy is in line with established practices from previous works, such as~\cite{GameGenX}. Instead of providing raw data, we furnish YouTube video IDs essential for accessing the content.

\item \textbf{Usage Rights:} The data we release is exclusively meant for research purposes. Any commercial use is not permitted under this agreement.

\item \textbf{Compliance with YouTube Policies:} Our data collection and sharing practices strictly adhere to YouTube’s data privacy and fair use policies. We ensure that user data and privacy rights are respected throughout the process.

\item \textbf{Data License:} The data is distributed under Creative Commons Attribution 4.0 International License (CC BY 4.0).
\end{itemize}
Furthermore, the DataDoP dataset is provided solely for informational purposes. The copyright for the original video content remains with the respective owners. All DataDoP videos are sourced from the internet and are not owned by our institution. We disclaim responsibility for the content and interpretation of these videos. In relation to the future open-source version, researchers must agree not to reproduce, duplicate, sell, trade, resell, or exploit any portion of the videos or derived data for commercial purposes, and refrain from copying, publishing, or distributing any part of the DataDoP dataset.

\input{Figure_sup/statistic_sup}
\subsection{Construction Details}
\label{sec:Construction_sup}
\noindent\textbf{Pre-processing.}
Our dataset construction involves a multi-stage curation process:
\begin{itemize}
\item \textbf{Shot Collection}: A curated collection of cinematographically significant films and documentaries forms the foundation of our dataset. Using PySceneDetect\footnote{https://github.com/Breakthrough/PySceneDetect}, we extract 43k initial shots through content-aware boundary detection. An optimized VSR pipeline\footnote{https://github.com/YaoFANGUK/video-subtitle-remover} is employed to eliminate textual overlays while maintaining visual integrity. To enhance processing speed and reduce misclassification, we focus the check area on the lower 1/5 of the frame. Finally, we merge this dataset with a publicly available subset of MovieShots~\cite{MovieShots} to further diversify stylistic elements.
\item \textbf{Quality Filtering}: We retained only shots with a duration between 10 and 20 seconds. Statistics can be seen in~\cref{fig_sup:length}.
Then we exclude sequences with static frames and low-light conditions.
For each shot, we calculate the pixel-wise similarity between all pairs of consecutive frames. The similarity between two frames \( F_1 \) and \( F_2 \) is defined as:
\[
S(F_1, F_2) = \frac{\sum_{i,j} \mathbb{I}(F_1(i,j) = F_2(i,j))}{H \times W},
\]
where \( F_1(i,j) \) and \( F_2(i,j) \) represent the pixel values at position \((i, j)\) in frames \( F_1 \) and \( F_2 \), respectively, and \( H \) and \( W \) are the height and width of the frames.
To identify static frames, we compute the average similarity \( \overline{S} \) between all consecutive frame pairs in the shot. Specifically, for a shot with \( N \) frames, the average similarity is calculated as:
\[
\overline{S} = \frac{1}{N-1} \sum_{k=1}^{N-1} S(F_k, F_{k+1}),
\]
where \( F_k \) and \( F_{k+1} \) are consecutive frames in the shot, and \( N-1 \) is the number of consecutive frame pairs.
If the average similarity \( \overline{S} \) exceeds a threshold (e.g., \( \overline{S} > 0.6 \)), the entire shot is considered static and excluded.

For each shot, the average brightness (mean gray value) for all frames is computed using the following formula:
\[
\overline{B} = \frac{1}{N \times H \times W} \sum_{k=1}^{N} \sum_{i=1}^{H} \sum_{j=1}^{W} F_k(i,j),
\]
where \( N \) is the total number of frames in the shot, \( H \) and \( W \) are the height and width of each frame, and \( F_k(i,j) \) represents the pixel value at position \((i,j)\) in frame \(k\).

If the average brightness of a shot is below a predefined threshold (e.g., \( \overline{B} < 15 \)), the shot is classified as too dark and excluded from the dataset.

\item \textbf{Semantic Filtering}: We developed an automated categorization pipeline using GPT-4o. Following the definitions in Sec. 3.2, shots are classified into categories. Leveraging GPT-4o~\cite{GPT4O}, we automate the categorization of shots into \textit{Static}, \textit{Free-Moving}, and \textit{Tracking}. Shots classified as \textit{Object/Scene-Centric}, which are common in multi-view datasets, are not considered in this study. We then discard the \textit{Static} and \textit{Tracking} shots. The detailed process and examples are shown in~\cref{fig_sup:gpt4o_cate}.

\input{Figure_sup/gpt4o_cate}
\end{itemize}

\noindent\textbf{Trajectory Extraction.}
We use MonST3R~\cite{MonST3R} to estimate the geometry of dynamic scenes, generating a time-varying dynamic point cloud along with per-frame camera poses and intrinsics in a feed-forward manner. This enables efficient video depth estimation and reconstruction~\cite{Voxurf}.
Camera trajectories, along with the corresponding depth maps, are extracted for further processing. These trajectories undergo a series of steps including cleaning, smoothing, interpolation, and standardization into fixed-length sequences, ensuring their suitability for subsequent training.

\begin{itemize} \item \textbf{Cleaning the Trajectories:}
To clean the camera trajectories, we first extract the camera translations from the transformation matrices and compute the velocities between consecutive frames. A threshold is determined based on the 95th percentile of the velocity distribution, with an outlier exclusion factor $\alpha$ (set to 18.0). Frames with velocities exceeding this threshold are discarded. The remaining valid frames are then grouped into consecutive segments, ensuring that each segment contains at least 5 frames.

\item \textbf{Smoothing the Trajectories:}
After the trajectories have been cleaned, a Kalman~\cite{kalman1960new} filter, based on a Constant Velocity model, is applied to smooth the valid frames within each segment. The smoothing process is performed using process and measurement noise standard deviations of 0.5 and 1.0, respectively. The smoothed segments are subsequently recombined with the original poses, resulting in a cleaned and smoothed camera trajectory. This smoothing step serves to reduce noise and enhance the stability and accuracy of the trajectory, facilitating more reliable analysis in subsequent stages.

\item \textbf{Interpolation into Fixed-Length Sequences:}
To ensure consistency across the trajectory data for downstream deep learning tasks, we standardize trajectories of varying lengths into fixed-length input sequences, addressing issues related to inconsistent time steps. First, spherical linear interpolation (SLERP)~\cite{shoemake1985animating} is applied to the rotational component, while the translational component is interpolated linearly, ensuring smooth transitions between frames. The interpolated data is then padded to a fixed length of 120 frames, ensuring uniform time steps across all trajectory samples. This process guarantees that the input sequences are consistent in length and temporal structure, providing stable and reliable training data for deep learning models. \end{itemize}
\input{Figure_sup/com}
\noindent\textbf{Motion Tagging.}
We present the distribution of translation and rotation combinations in~\cref{fig_sup:com}. As shown, simpler motion combinations are more frequent, but motion tags still exhibit high diversity and complexity.

\input{Figure_sup/gpt4o_caption}
\noindent\textbf{Caption Generation.}
In~\cref{fig_sup:gpt4o_caption}, we present the specific prompts and cases for generating two types of captions.

\section{GenDoP Method}
\subsection{Camera Trajectory Tokenization Details}
\label{sec:Method_sup}
The camera trajectory tokenization process converts continuous camera parameters into discrete tokens. For the normalized parameters $\mathbf{R}_i^{\mathrm{norm}}$ (rotation) and $\mathbf{t}_i^{\mathrm{norm}}$ (translation) obtained after canonical normalization, the rotation matrix $\mathbf{R}_i^{\mathrm{norm}}$ is converted into a unit quaternion $\mathbf{q} = (r_1', r_2', r_3', r_4')$, and the translation values $\mathbf{t}_i^{\mathrm{norm}} = (t_1', t_2', t_3')$ are processed. This results in the combined vector $(r_1', r_2', r_3', r_4', t_1', t_2', t_3')$. Subsequently, the focal lengths $f_x$, $f_y$, and the scale factor $s$ are also acquired. The process involves three key stages:
\begin{itemize}
\item \textbf{Rotation and Translation Normalization}: The rotation and translation components are tokenized as follows:

\[
r_k = \frac{r_k' + 1}{2}, \quad k \in \{1, \dots, 4\},
\]

\[
t_k = \frac{t_k' + 1}{2}, \quad k \in \{1, \dots, 3\}.
\]

This normalization maps values from the range $[-1,1]$ to $[0,1]$, while preserving the constraints on both rotation and translation.
 
\item \textbf{Focal Length Adaptation}:
Normalize focal lengths $(f_x, f_y)$ relative to the principal points $(c_x, c_y)$:
\[
            f_1 = \frac{f_x}{10c_x},\\
            f_2 = \frac{f_y}{10c_y}.
\]
    Here, $c_x$ and $c_y$ typically represent half the image dimensions. The factor of 10 ensures that the focal length values remain within the range $(0,1)$, accommodating typical focal lengths.

\item \textbf{Scale Parameter Transformation}:
        Apply logarithmic compression to the scale factor $s$: \[ s = \frac{\log_{10}(s') + 2}{4}.\]
        This transformation enables a linear representation of multiplicative scale changes across three orders of magnitude ($0.01 \leq s' \leq 100$).

\item \textbf{Parameter Clamping and Discretization}:
    All normalized parameters are clamped to the range $[0,1]$ before discretization into $N$ bins: \[ p^{token} = \lfloor p \cdot N \rfloor, \quad \forall p \in {r_1,\dots, t_3, f_1, f_2, s}. \ \] This process generates a compact 10-dimensional token that preserves the relative geometric relationships between parameters. The hyperparameter $N$ controls the trade-off between quantization error and codebook size.
\end{itemize}

\input{Tables/ablation_sup}
\section{Experiments}
\subsection{Experiments Setting Details}
\label{sec:Experiments_sup}
We train GenDoP with a batch size of 16 using the AdamW optimizer, with a learning rate of 1e-5, $(\beta_1, \beta_2) = (0.9, 0.95)$, and a weight decay of 0.01. The KL loss weight is set to 1e-8. We use a gradient accumulation step of 1. Training is performed using bfloat16 mixed precision. The model converges with the best results at the 100th epoch.

\subsection{Additional Qualitative Results}
In the supplementary video, we present additional cases with text prompts randomly generated by the LLM model~\cite{GPT4O}. These text prompts have never been seen in the training set, creating a certain gap compared to the captions in our dataset. However, as shown in the video, despite the differences between the generated prompts and the training data, our model is still able to generate precise, high-quality, complex, and artistic trajectories. 

Furthermore, we use TrajectoryCrafter~\cite{TrajectoryCrafter} to showcase how our trajectory generation method can be applied to camera-controlled video generation. This allows for the creation of videos that align with the camera descriptions provided, ensuring the generated video sequences match the specific visual and motion criteria described by the camera control inputs.

\subsection{Additional Ablation Studies}
We conduct ablation experiments on several hyperparameters, as shown in~\cref{tab:ablation_sup}, including the number of discrete bins, trajectory length, and model size. These parameters correspond to the discrete bin size \( B \), the trajectory length, and the model size (as detailed in Sec. 5.1). Specifically, for the \texttt{small} size, the latent dimension is \( L = 512 \), with the backbone OPT Transformer consisting of 8 layers and 8 attention heads per layer. For the \texttt{base} size, the latent dimension is \( L = 1024 \), with 12 layers and 12 attention heads. For the \texttt{large} size, the latent dimension is \( L = 1536 \), with 16 layers and 24 attention heads.

The results indicate that optimal performance is achieved when the number of discrete bins is set to 256, the trajectory length to 30, and the model size to \texttt{base}. Notably, when the model size is set to \textit{large}, although the performance in Text-texttt Alignment decreases, the Trajectory Quality improves. We speculate that this may be due to the larger model's tendency to overfit, learning better trajectory quality while failing to follow the text instructions effectively.

%% file: Figure_sup/statistic_sup.tex
\begin{figure*}
  \centering
  \begin{subfigure}{0.45\linewidth}
\includegraphics[width=1\linewidth]{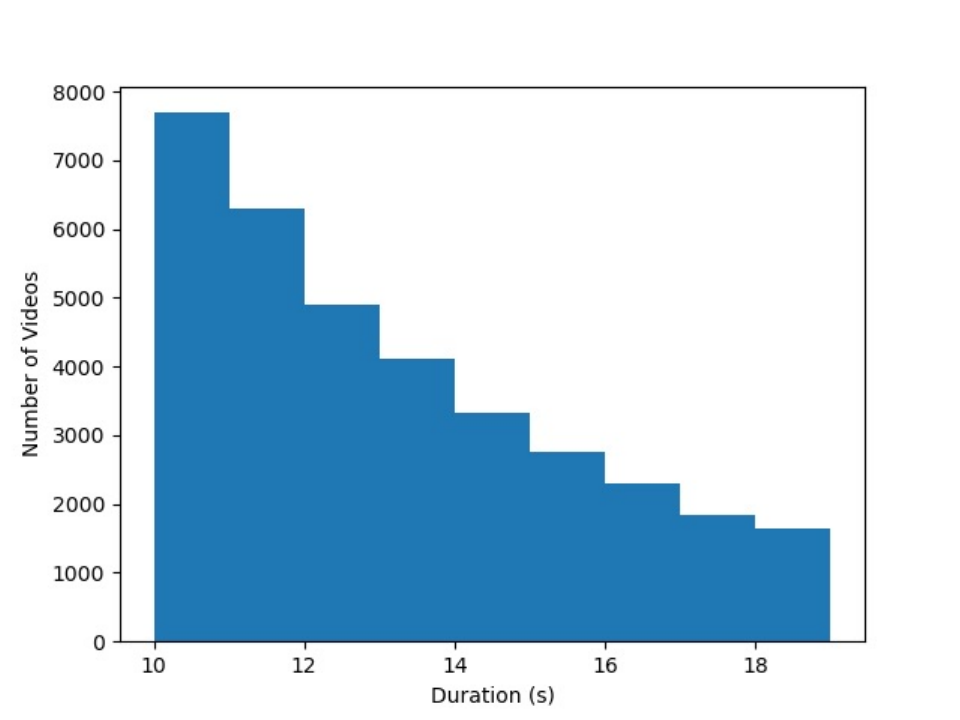}
    \caption{Shot Length.}
    \label{fig_sup:length}
  \end{subfigure}
    \hfill
    \begin{subfigure}{0.45\linewidth}
    \includegraphics[width=1\linewidth]{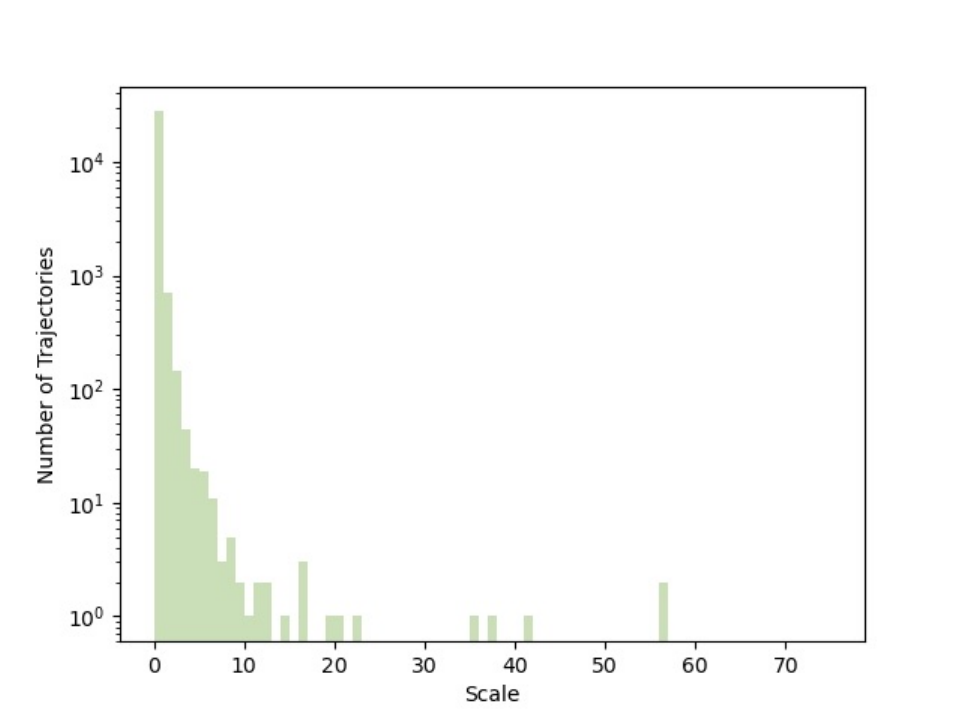}
    \caption{Trajectory Scale.}
    \label{fig_sup:scale}
  \end{subfigure}
  
  \caption{\textbf{Dataset Statistics} in terms of video shot length and trajectory scale.}
  \label{fig_sup:Statistics}
\end{figure*}

%% file: Figure_sup/gpt4o_cate.tex
\begin{figure*}
  \centering
    \includegraphics[width=1\linewidth]{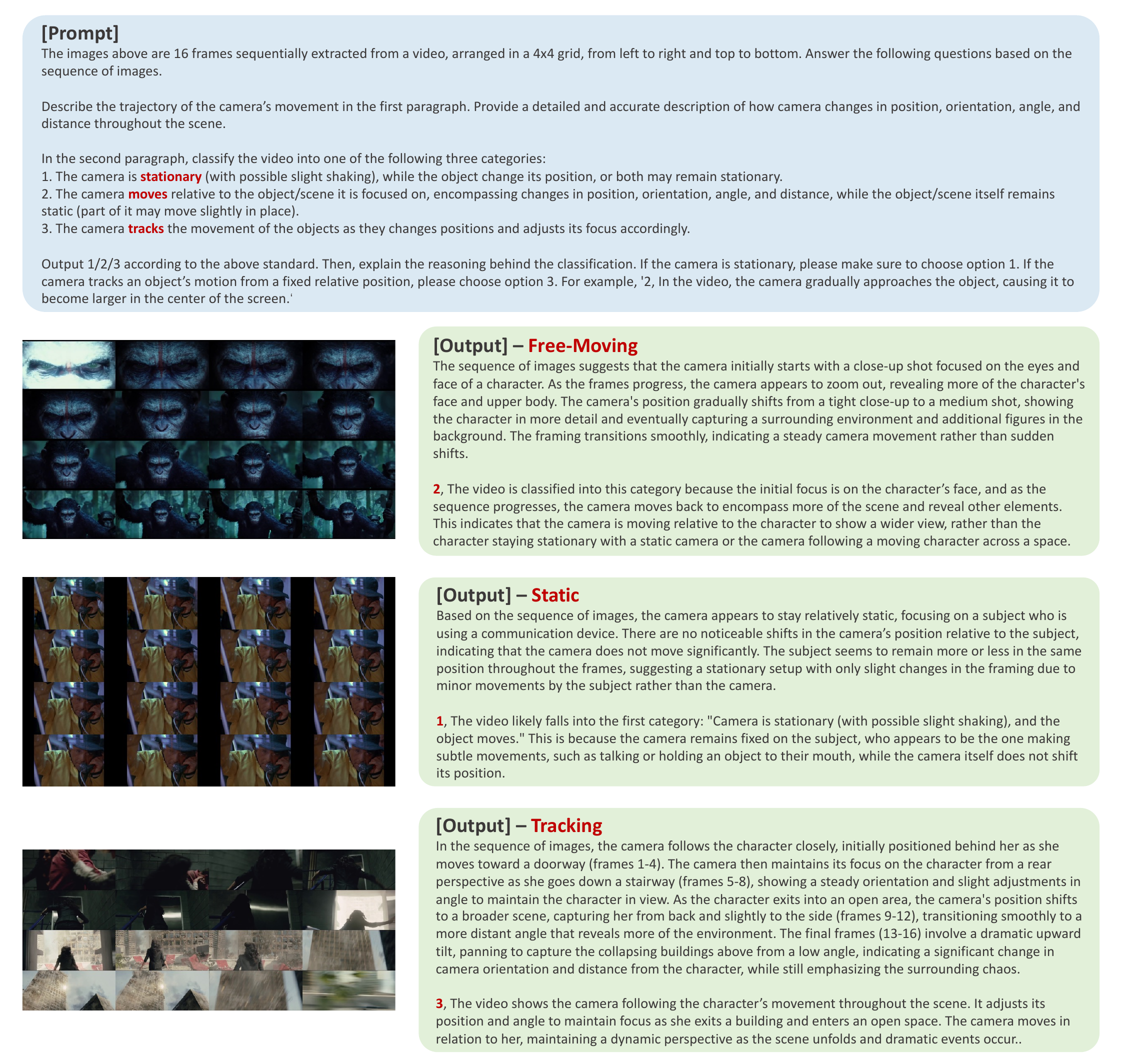}
\caption{\textbf{Semantic Filtering.} Following the definitions in Sec. 3.2, shots are classified. Leveraging GPT-4o~\cite{GPT4O}, we automate shot categorization into \textit{Static}, \textit{Free-Moving}, and \textit{Tracking}. Shots categorized as \textit{Object/Scene-Centric}, common in multi-view datasets, are not considered in films.}
  \label{fig_sup:gpt4o_cate}
\end{figure*}

%% file: Figure_sup/com.tex
\begin{figure*}
  \centering
    \includegraphics[width=1\linewidth]{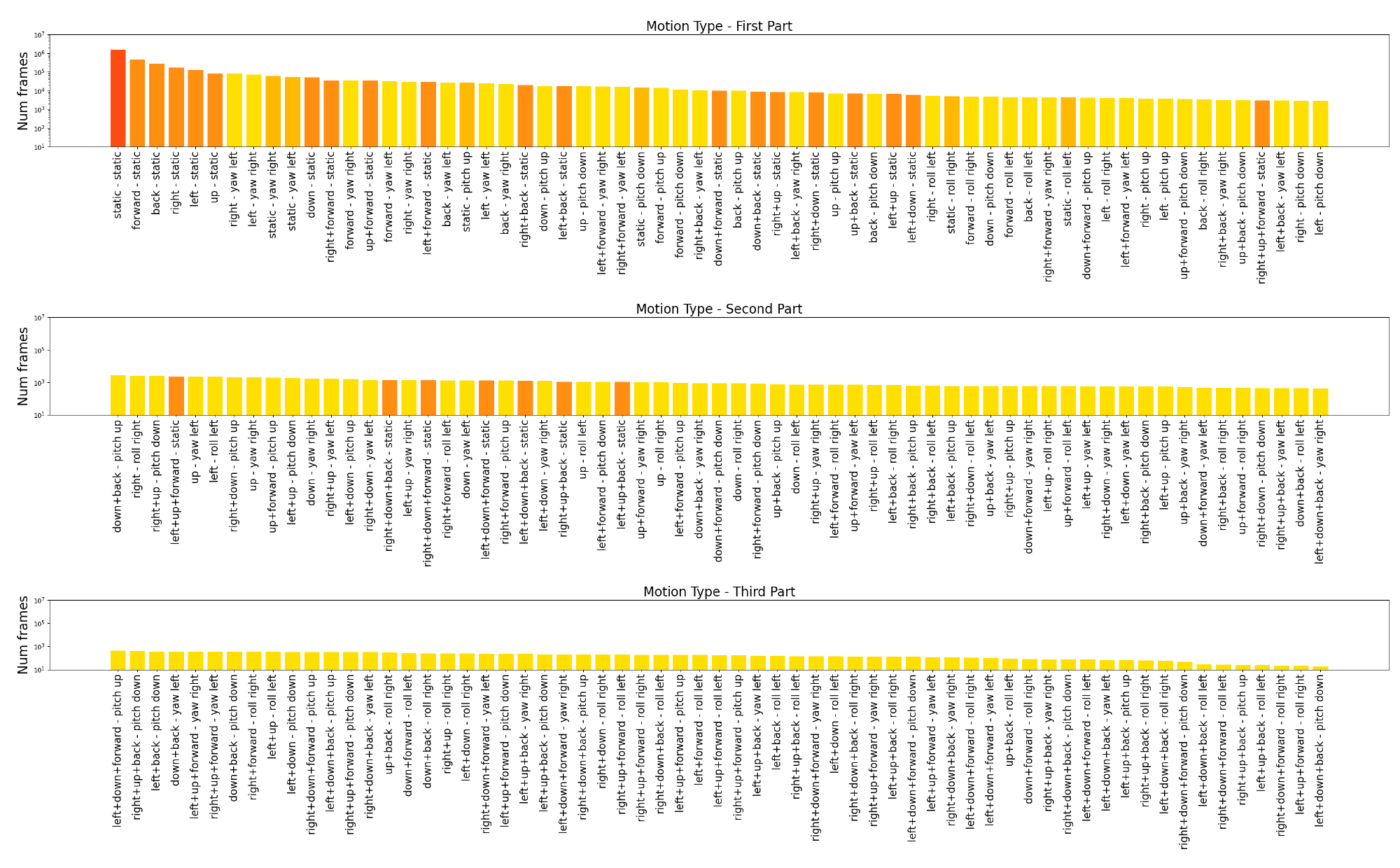}
\caption{\textbf{Tag Distribution.} The distribution of Translation and Rotation combinations is shown in the figure. Different tag modes are represented by shades of yellow, ranging from deep to light: Static, Translation only, Rotation only, and both Translation and Rotation.}
  \label{fig_sup:com}
\end{figure*}

%% file: Figure_sup/gpt4o_caption.tex
\begin{figure*}
  \centering
    \includegraphics[width=1\linewidth]{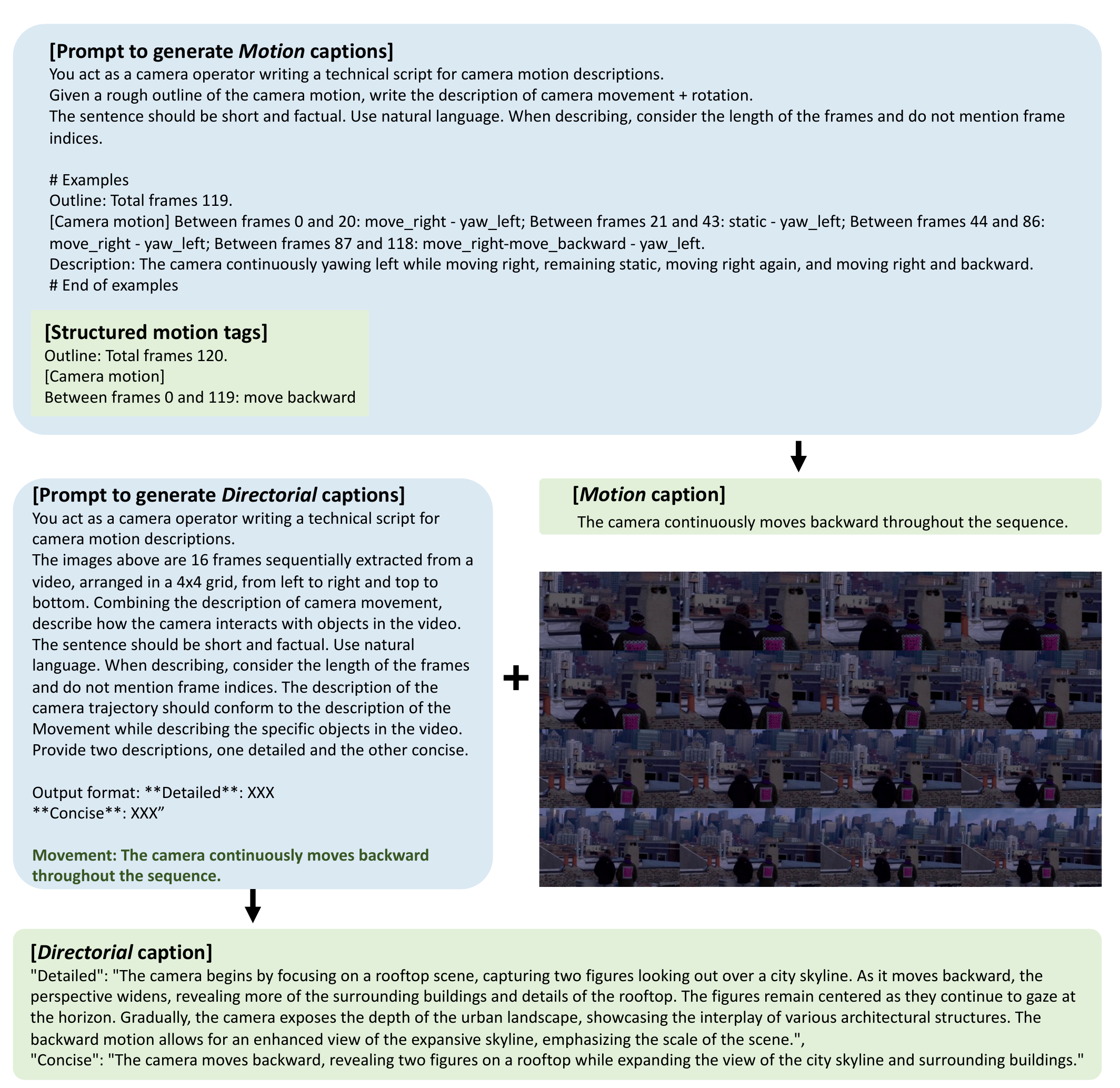}
\caption{\textbf{Caption Generation.} We structure the motion tags by incorporating context, instructions, constraints, and examples, and then leverage GPT-4o to generate \textbf{Motion} captions that describe the camera motion alone. Next, we extract 16 evenly spaced frames from the shots to create a $4\times4$ grid, prompting GPT-4o to consider both the previous caption and the image sequence. This enables GPT-4o to generate \textbf{Directorial} captions that describe the camera movement, the interaction between the camera and scene, and the directorial intent.}

  \label{fig_sup:gpt4o_caption}
\end{figure*}

%% file: Tables/ablation_sup.tex
\begin{table*}[h]
% \small
% \scriptsize
\centering
\small
\setlength{\tabcolsep}{20pt}
\renewcommand{\arraystretch}{0.95}
% \fontsize{6.8}{9}\selectfont
\begin{tabular}{lc|cc|cc}
\toprule
\multicolumn{2}{c}{\multirow{2}{*}{\textbf{Ablation}}} & \multicolumn{2}{c}{\textbf{Text-Trajectory Alignment}}                           & \multicolumn{2}{c}{\textbf{Trajectory Quality}}                                \\
\multicolumn{2}{c}{}                                   & \multicolumn{1}{c}{\textbf{F1-Score}$\uparrow$} & \multicolumn{1}{c}{\textbf{CLaTr-Score}$\uparrow$} & \multicolumn{1}{c}{\textbf{Coverage}$\uparrow$} & \multicolumn{1}{c}{\textbf{CLaTr-FID}$\downarrow$} \\
\midrule
\multirow{5}{*}{Discrete bins}         & 64            & 0.394                                 & 33.594                                   & 0.751                                 & 49.854                                 \\
                                       & 128           & \textbf{0.409}                        & 35.824                                   & 0.851                                 & 24.748                                 \\
                                       & 256           & 0.400                                 & \textbf{36.179}                          & 0.872                                 & \textbf{22.714}                        \\
                                       & 512           & 0.391                                 & 35.201                                   & 0.882                                 & 23.633                                 \\
                                       & 1024          & 0.393                                 & 34.277                                   & \textbf{0.884}                        & 24.979                                 \\
\midrule
\multirow{3}{*}{Traj length}           & 15            & 0.398                                 & 34.576                                   & 0.863                                 & \textbf{22.238}                        \\
                                       & 30            & \textbf{0.400}                        & \textbf{36.179}                          & \textbf{0.872}                        & 22.714                                 \\
                                       & 60            & 0.393                                 & 34.523                                   & 0.864                                 & 26.307 \\
\midrule  
\multirow{3}{*}{Model size}            & small          & 0.389                                 & 32.868                                   & 0.880                                 & 25.604                                 \\
                                       & base         & \textbf{0.400}                        & \textbf{36.179}                          & 0.872                                 & 22.714                                 \\
                                       & large         & 0.398                                 & 33.843                                   & \textbf{0.888}                        & \textbf{20.474} \\  
                                       \bottomrule

\end{tabular}
\caption{\textbf{Ablation Study on Hyperparameters.} We conduct ablation experiments on several hyperparameters, including the number of discrete bins, trajectory length, and model size. These parameters correspond to the discrete bin size $B$, the trajectory length $N$, and the model size (as detailed in Sec. 5.1). The results show that the optimal performance is achieved when the number of discrete bins is set to 256, the trajectory length to 30, and the model size to \texttt{base}.}
% \vspace{-20pt}
\label{tab:ablation_sup}
\end{table*}